\documentclass[conference]{IEEEtran}
\usepackage{amsmath,amssymb}
\usepackage{booktabs}
\usepackage{array}
\usepackage{graphicx}
\usepackage{xcolor}
\usepackage{tikz}
\usetikzlibrary{arrows.meta,positioning,fit,backgrounds,calc}
\usepackage{float}
\usepackage{url}
\usepackage[hidelinks]{hyperref}
\usepackage{cite}
\newcommand{\figref}[1]{Fig.~\ref{#1}}
\newcommand{\tabref}[1]{Table~\ref{#1}}
\newcommand{\kw}[1]{#1}
\newcolumntype{P}[1]{>{\raggedright\arraybackslash}p{#1}}
\newcommand{\Rsq}{R^2}
\newcommand{\RWKVself}{RWKV self-reconstruction control}
\newcommand{\Pythiaself}{Pythia self-reconstruction control}
\newcommand{\RtoP}{RWKV$\to$\allowbreak Pythia}
\newcommand{\PtoR}{Pythia$\to$\allowbreak RWKV}
\newcommand{\RtoPL}[1]{\RtoP{} ($L{=}#1$)}
\newcommand{\PtoRL}[1]{\PtoR{} ($L{=}#1$)}
\newcommand{\hfrepo}[2]{%
  \href{#1}{%
    \texttt{https://huggingface.co/}\allowbreak\texttt{Katagiri-Hoshino-Lab/}\allowbreak\texttt{#2}}}

\begin{document}

\title{NinaXander: Feasibility and Limits of Composing Frozen Language Models Across Architecture Families via a Shared Latent Space}

\author{\IEEEauthorblockN{Takanori Kotama and Shun-ichiro Hayashi}
\IEEEauthorblockA{Graduate School of Informatics\\Nagoya University}
\and
\IEEEauthorblockN{Daichi Mukunoki, Tetsuya Hoshino and Takahiro Katagiri}
\IEEEauthorblockA{Information Technology Center\\Nagoya University}}

\maketitle

\begin{abstract}
In this paper we propose \kw{NinaXander}, a series of composed language models obtained by connecting layers of frozen language
models from different architecture families with a single trained shared-latent adapter.
A composed model runs the first layers of one model, converts the resulting intermediate representation once with the adapter,
and then runs the remaining layers of the other model.
Once the adapter is trained, several composed models that connect at different layers are obtained without retraining.
Using the recurrent RWKV-4-Raven-7B and the Transformer-based Tulu-Pythia-6.9b, abbreviated as RWKV and Pythia, this study examines whether frozen models from
different families can be recombined post hoc.
The composed models answered multiple-choice questions, and those whose generations we examined produced syntactically well-formed text.
The configuration that combines the first $5$ layers of Pythia with the
remaining $27$ layers of RWKV reduced the Transformer key-value (KV) cache by $84.4\%$ with accuracy not significantly
different from that of RWKV alone. In multiple-choice accuracy, however, no composed model matched the parent model Pythia, and language-modeling
performance decreased sharply on WikiText, a corpus of Wikipedia articles outside the training domain. The correspondence between intermediate representations was also
obtained in one favorable case, with a shared tokenizer, the same depth, and the same hidden width, and does not show that the models share
a general semantic space.
\end{abstract}

\section{Introduction}\label{sec:intro}
Pretraining a language model requires an enormous amount of computation, yet its result remains only as weights fixed to a
particular architecture. The Transformer, which most current language models use, keeps the keys and values of past tokens as a
key-value (KV) cache to compute self-attention, and the size of this cache grows in proportion to the context length. Existing ways
of changing inference-time characteristics through the architecture, for example to reduce the KV cache, either pretrain a new
hybrid model that incorporates recurrent layers or replace some layers of an existing model with a lighter mechanism and retrain it.
Both approaches retrain weights; neither recombines trained blocks as frozen components.

In this paper, a family is a group of models that share the same kind of architecture, such as Transformers or RWKV (Receptance Weighted Key Value), which computes linear attention in a recurrent form and so runs in time linear in the
sequence length.
Studies that connect frozen language models to one another have been reported, but in these studies either both connected models are
Transformers or the whole composed model is fine-tuned, and recombining trained checkpoints across families as frozen components has not
yet been examined.

This study examines whether the intermediate representations of frozen models from different families can be converted into each
other by a small trained network, called the adapter below, so that the first layers of one model can be connected to the remaining
layers of another. We call a model composed in this way a chimera model and the original models used for the composition parent
models, and we name the series of chimera models built with a single adapter
\kw{NinaXander}\footnote{The name NinaXander comes from a chimera in Hiromu Arakawa's manga \emph{Fullmetal Alchemist}. The
composition works but falls short of what it was made from, which corresponds to the results of this study.}.
Once the adapter is trained, several chimera models that connect at different layers are obtained without retraining.
The parent models are RWKV-4-Raven-7B, a version of RWKV further trained to follow instructions, and
the Transformer-based Tulu-Pythia-6.9b. The two models satisfy the favorable conditions of sharing a tokenizer, which splits text into token sequences, and
of having the same hidden width and the same number of layers. Below, the two models are abbreviated as RWKV and Pythia, respectively.

The main purpose of this study is not to build a practical high-performance model but to clarify whether frozen models from
different families can be connected by converting their intermediate representations. In our experiments the chimera models
answered multiple-choice questions, and those whose generations we examined produced syntactically well-formed text. We also obtained a configuration that reduces the
Transformer layers to $5$, decreasing the KV cache size by $84.4\%$, with accuracy not significantly different from that of RWKV alone.
On the other hand, no configuration exceeded the stronger parent model, and RWKV alone, which has no KV cache, was significantly
exceeded in accuracy by only one chimera model, on the science question set SciQ. The contributions of this study are the construction and evaluation of chimera models that connect frozen
language models from different families in both directions with a single adapter; an analysis of the factors that limit the
performance of the connection, through a comparison with reconstruction within the same model and an intervention inside the
adapter; and measurements of accuracy, KV cache size, and performance on text from domains not used to train the adapter.

\section{Background and Related Work}\label{sec:related}
\subsection{Residual Stream and Inference Cost}
An autoregressive language model updates the representation at each token position, called the residual stream, layer by layer.
Block $\ell$, counted from $0$, of a model with hidden width $d$ receives the input $h^{\ell-1}$ and outputs $h^{\ell}=h^{\ell-1}+f_\ell(h^{\ell-1})$,
where $h^{-1}$ is the token embedding.
The function $f_\ell$ performs sequence mixing, which mixes information across tokens, and channel mixing, which transforms features
within each token position. Even when two models have the same hidden width and depth, the $h^{\ell}$ of one model lies in a coordinate
system determined by that model's training, and passing it without conversion to block $\ell{+}1$ of the other model does not match
the distribution that the receiving weights expect. Whether two models can be connected at an intermediate layer therefore reduces
to whether the residuals of the two families can be converted into each other.

The two families also differ in how their inference cost grows. The self-attention of Transformers weights tokens by their
similarity normalized with the softmax function, and because it keeps the keys and values of all past tokens it needs a cache of
size $O(T)$ for a context length $T$. A layer that can run as linear attention or in an equivalent recurrent form keeps only a
fixed-size state, which is $O(1)$. Which of the two connected models is placed on the input side changes how many Transformer
layers remain and how much cache must be kept.

\subsection{Model Stitching}
Lenc and Vedaldi~\cite{10.1007/s11263-018-1098-y} introduced connecting layers of $2$ frozen models through a trainable stitching
layer as a way to measure the equivalence of representations. For two convolutional neural networks (CNNs), they built a
``Franken-CNN'' that connects the lower layers of one network to the upper layers of the other through a linear stitching layer, and
evaluated it with the loss of the original task. The pairs they studied include networks trained on different data, namely the
large-scale image classification dataset ImageNet and the scene recognition dataset MIT Places built by the Massachusetts Institute
of Technology (MIT). They also include pairs of CNNs with different architectures: the 8-layer AlexNet, the 16-layer VGG-16 proposed by the
Visual Geometry Group of the University of Oxford, and the 50-layer ResNet-50 with residual connections.
Bansal et al.~\cite{3540261.3540279} revisited stitching as a means of comparing representations. They stitched models of the same
architecture that differ in random seed, amount of training, and width, and also stitched across different objectives, namely
supervised learning and self-supervised learning, which trains on pseudo-labels constructed from the data itself. Stitching across
different architectures, however, remained an open problem. Csisz\'arik et al.~\cite{NEURIPS2021_2cb274e6} matched the intermediate
representations of two trained networks with a single affine stitching layer and proposed a framework that evaluates their agreement
not only by similarity in representation space but also functionally, by the task performance of the stitched network. They further
showed that existing measures of representational similarity do not necessarily reflect task performance. Following this position,
we evaluate separately the degree to which representations agree, called alignment below, and the degree to which one representation
can be recovered from the other, called decodability below and measured in this paper by the read-out accuracy
(\S\ref{sec:common-mode}). All of these studies target image recognition models. This study differs in that it connects two
different families, recurrent neural networks (RNNs) and Transformers, through a common latent space in which the representation is
converted once at an intermediate layer.
\subsection{Composing Frozen Large Language Models}
Recent studies compose frozen large language models (LLMs) through trainable connection layers.
StitchLLM~\cite{hu-etal-2025-stitchllm} connects blocks of Transformers of different sizes through stitching layers and uses a
lightweight routing mechanism to decide how much computation to allocate to each query. BTS (Branch-Train-Stitch)~\cite{zhang-etal-2025-bts}
places several frozen expert models, derived from the same seed model for specific domains, alongside the frozen seed model, which serves as the hub, and
inserts Experts-into-Hub and Hub-into-Experts stitching layers alternately every few layers to exchange hidden states in both
directions. CALM (Composition to Augment Language Models)~\cite{bansal2024llm} fuses the intermediate representations of two frozen
Transformers through cross-attention, which uses the representation of one model as queries to attend to the representation of the
other. Manticore~\cite{roberts2025pretrained} runs block sequences of a Transformer and of Mamba, a selective state-space model, in
parallel with projections, and mixes their outputs with learned weights that are non-negative and sum to 1. This hybrid model is
fine-tuned as a whole, and its embedding layer and its language-model head (LM head), which outputs probabilities over the vocabulary,
are newly initialized.

Other studies convert activations directly between frozen models. Moschella et al.~\cite{moschella2023relative} showed that
relative representations, which re-express each representation by its similarity to a small set of reference samples called anchors,
remove the effect of the transformation between independently trained latent spaces, so that models can be stitched without training a converter.
Maiorca et al.~\cite{3666122.3668540} estimate a map between two latent spaces in closed form from a small number of semantic
correspondences. As low-capacity direct maps corresponding to these methods, this study uses affine maps for comparison
(\S\ref{sec:linear}). Because our affine maps are fitted directly with corresponding residuals as targets, the comparison is more
favorable to them than to relative representations, which do not use paired data for training.

Model Alignment Search~\cite{grant2025model} maps activations in both directions between a frozen recurrent network and a frozen
Transformer, but it is a method for measuring representational similarity and does not build a composed model.
Chen et al.~\cite{chen2025transferring} map the residual streams of two frozen language models in both directions with a single
affine layer, and evaluate, down to the cross-entropy computed from next-token prediction probabilities, composed models that connect the
first layers of one model to the remaining layers of the other. Their pairs, however, are Transformers of one model series sharing
a tokenizer; the future work they name is validation across model series, and they do not discuss different families.

The Bicameral Model~\cite{flamant2026bicameralmodelbidirectionalhiddenstate}, a preprint, couples two frozen language models through
bidirectional conversion networks, but each model runs in parallel while executing all of its own layers. GoldFinch~\cite{goldstein2024goldfinchhighperformancerwkvtransformer},
also a preprint, stacks a Transformer on a structure that extends RWKV-6 (Finch), an improved version of RWKV-4, and
compresses the KV cache by $756$--$2550$ times, but it is trained from an initial state rather than combining pretrained parent
models. Oozeer et al.~\cite{oozeer2025activation} learn activation maps between frozen LLMs and transfer steering vectors that control
tendencies of the output, mainly between smaller and larger models of one series, using them to remove backdoors planted so that specific inputs trigger undesired outputs and to
disable the refusal of harmful instructions. To validate a
map, they replace the activations of one target layer with mapped ones and generate text, and they use affine maps as a control, so
their work is close to ours; each map, however, serves one layer pair, and both models are Transformers.

\subsection{Representation Similarity}
Standard measures of representational similarity between networks are centered kernel alignment (CKA)~\cite{pmlr-v97-kornblith19a}
and two extensions of canonical correlation analysis (CCA), singular vector CCA (SVCCA) and projection weighted CCA
(PWCCA)~\cite{3295222.3295356,3327345.3327475}.
Of these, we use linear CKA, together with the fitted $R^2$ of per-layer linear maps, as auxiliary indicators to examine the
correspondence across all layers, and focus on functional stitching
through reconstruction. On the universality of representations across families, Wang et al.~\cite{wang2025towards} compared
Transformers and Mamba. They extracted the features of both models with sparse autoencoders, which are trained to express the input
as a combination of a small number of features, and reported that most features are similar. They further showed that the structure
of induction circuits, which predict the next token by referring to repeated patterns in the context, is also similar. Their subject,
however, is Mamba, not the RWKV family.
\subsection{Model Merging and Knowledge Fusion}
Methods that merge models in weight space include model soups~\cite{pmlr-v162-wortsman22a}, which average the weights of several
models fine-tuned from the same pretrained model; TIES-Merging~\cite{3666122.3666432}, which merges task-specific weight differences
after resolving sign conflicts; and evolutionary model merge~\cite{Akiba2025}, which searches for merging recipes by evolutionary
computation. These methods assume identical or compatible weights. FuseLLM~\cite{wan2024knowledge}, which fuses the output
distributions of several models, does not need compatible weights, but it needs a procedure that aligns tokens between the
vocabularies of different tokenizers and additional training of the target model. This study merges no weights and recombines
frozen blocks by converting in activation space, which makes it complementary to these methods.

\subsection{Position of This Study}
Among the studies above, those that compose language models either connect two Transformers, fine-tune the composed model as in
Manticore, or use at most one pretrained parent model. A stitching scheme with a single connection point, connections in both directions, and frozen parent
models has already been demonstrated by Chen et al.~\cite{chen2025transferring} for pairs of Transformers within one model series that share a
tokenizer, and this study does not claim the scheme itself as a new contribution. What this study addresses is the case of
different families, which Chen et al. did not examine: we test whether the same stitching holds between a recurrent model based on linear
attention and a Transformer. To our knowledge, no study has composed these two families by connecting them at a single point with both
parent models and the LM heads frozen and evaluated the result as a generator. We also report how much the composition reduces the
Transformer KV cache.

\section{Method}\label{sec:method}
\subsection{Target Models}
We use RWKV-4-Raven-7B and Tulu-Pythia-6.9b, instruction-following models tuned to respond according to user
instructions\footnote{The experiment code, raw evaluation data, and reproduction procedures are available at \url{https://github.com/Katagiri-Hoshino-Lab/NinaXander}.}.
Below, the former is denoted model $A$ and the latter model $B$. These symbols are used in the notation for residuals in
\S\ref{sec:notation} and for read-outs in \S\ref{sec:config}. Both models have $32$ blocks and a hidden width of $d{=}4096$, and
they share the tokenizer of GPT-NeoX-20B, a language model developed by the non-profit research group EleutherAI.
The same input is therefore converted into the same token sequence, and the residuals at each position can be paired directly.
Because the depths are equal, the $j$th blocks of the two models can be paired, and because the hidden widths are equal, the vectors
at the connected layer have the same dimension. These three conditions were chosen deliberately to make the connection easier, and
pairs of models whose tokenizers, depths, or hidden widths differ are outside the scope of this study.

The base models of both were pretrained on The Pile~\cite{gao2020pile}, an $825$ GiB English corpus that EleutherAI built from $22$ kinds of data, including academic papers, web documents, books,
and source code. Although the base corpus is shared, the composition of the data and the prompt formats used for additional training on pairs of
instructions and responses, called instruction tuning below, differ between the two models.

The RWKV architecture~\cite{peng-etal-2023-rwkv} replaces softmax dot-product attention with linear-time sequence mixing by the WKV operator, which
averages the values of the tokens up to the current one with weights set by their keys and decayed over time, and it can run the same computation either in parallel
form or in recurrent form. The Raven model used here is an instruction-following version of RWKV-4 developed by the RWKV project and
released by its developer BlinkDL. RWKV-4-Pile-7B, pretrained on The Pile itself, is further trained on data that
include Alpaca~\cite{alpaca}, the instruction-response data released by Stanford University, the instruction data Guanaco and GPT4All,
Code-Alpaca, and ShareGPT.
Code-Alpaca is synthetic data of about $20{,}000$ instructions and responses centered on code generation, editing, and optimization
~\cite{codealpaca}.
ShareGPT collects multi-turn conversations with ChatGPT, the dialogue service of OpenAI, shared by its users
~\cite{lmsys-fastchat}.
We convert the RWKV-4-Raven-7B released by BlinkDL into the format used on the model-sharing platform Hugging Face and re-save it in 16-bit
floating point (FP16)\footnote{BlinkDL, RWKV-4 Raven model repository,
\url{https://huggingface.co/BlinkDL/rwkv-4-raven}.}.
The weights we use have $7{,}392{,}649{,}216$ parameters, $32$ blocks, a hidden width of $4096$, and an intermediate width of $16384$ in channel mixing.
The converted configuration sets \texttt{context\_length} to $1024$. This value is defined as the maximum sequence length handled in
a single forward pass, and in recurrent form the model can propagate a state whose size does not depend on the context
length\footnote{Hugging Face Transformers, RWKV documentation,
\url{https://huggingface.co/docs/transformers/model_doc/rwkv}.}. The file name and revision of the source weight file,
however, were not recorded, so the context length used to train the source checkpoint cannot be identified. Being recurrent also
does not guarantee language performance on sequences longer than the training or configured length.

The Pythia series~\cite{3618408.3618510} consists of Transformer models that EleutherAI developed for research and trained with its
GPT-NeoX library. Tulu-Pythia-6.9b is Pythia-6.9b with all of its parameters fine-tuned by the Allen Institute for AI (AI2), a non-profit
research institute, on the mixed instruction data Tulu~\cite{3666122.3669390}. The mixture contains FLAN V2, which converts many
natural language processing tasks into instruction format and whose name stands for Finetuned Language Net; Chain-of-Thought, which
collects answers that include reasoning steps; Dolly, instruction data written by employees of Databricks; OpenAssistant 1,
conversation data collected by crowdsourcing; GPT4-Alpaca, in which GPT-4, a large language model of OpenAI, responds to the Alpaca
instructions; and Code-Alpaca and ShareGPT\footnote{Allen Institute for AI,
open-instruct-pythia-6.9b-tulu model card,
\url{https://huggingface.co/allenai/open-instruct-pythia-6.9b-tulu}.}.
The weights we use have $6{,}856{,}040{,}448$ parameters, $32$ blocks, a hidden width of $4096$,
and $32$ attention heads. The maximum number of positions in the configuration is $2048$\footnote{The released configuration file is
\url{https://huggingface.co/allenai/open-instruct-pythia-6.9b-tulu/blob/main/config.json}, which sets
\texttt{max\_position\_embeddings} to $2048$.}.
According to their publishers, the instruction-tuning data of both models include datasets named Code-Alpaca and ShareGPT, and
Tulu's GPT4-Alpaca uses the same instructions as Raven's Alpaca. Raven's publisher does not state versions, so the extent of the
overlap cannot be determined, but the instruction data of the two models are not independent.
In this paper, block and layer are used as synonyms.

\subsection{Residuals and Notation}\label{sec:notation}
In this paper a residual does not mean an error; it is the hidden vector passed from each block to the next. Connecting two models
means converting this vector from the coordinate system of one model into that of the other and passing it on.
We write $h_M^j[t]\in\mathbb R^{4096}$ for the unstandardized residual that block $j$ of model $M\in\{A,B\}$ outputs at token position $t$.
It corresponds to $h^{\ell}$ in \S\ref{sec:related}, and the block index is $j\in\{0,\ldots,31\}$.
We feed the same token sequence to both models and take $h_A^j[t]$ and $h_B^j[t]$ as one training example.

The scale of the residuals differs between models and layers, so we compute the mean and standard deviation of each dimension from
the training data of each layer. $\mathrm{std}_M^j$ standardizes a residual with these statistics, and $\mathrm{unstd}_M^j$ is the
inverse operation that restores the original scale. The standardized residual is written $x_M^j[t]=\mathrm{std}_M^j(h_M^j[t])$. In the
training objective below the layer index $j$ and the token position $t$ are omitted for brevity, and we write $x_A,x_B$.

\subsection{Shared-Latent Adapter}\label{sec:adapter}
Our adapter maps the intermediate representations of the two parent models into a common vector space, called the shared latent
space below, and recovers the intermediate representation of the other model from this space. We call this adapter the shared-latent
adapter. It has one encoder and one decoder for each model. $E_A,E_B$ are encoders that map standardized residuals into the shared
latent space, and $D_A,D_B$ are decoders that recover each model's standardized residuals from the shared latent space. The subscripts
$A,B$ indicate the model to which each map belongs.
$E_A,E_B,D_A,D_B$ form a single adapter shared by all $32$ layers, with no conditioning on the layer index. Differences between layers
are handled only by the statistics of standardization and unstandardization.
$z_A,z_B\in\mathbb R^{d_z}$ are the two latent vectors obtained at the same token position, and the latent width is $d_z{=}4096$.
We call recovering a residual from the shared latent space a read-out. A read-out that uses the encoder and decoder of the same model
is a self-reconstruction, and a read-out that combines the encoder of one model with the decoder of the other is a cross-family
read-out.
With $\lambda\geq0$ the hyperparameter that weights the latent alignment error against the two self-reconstruction errors, the
training objective is defined as follows.
\begin{equation}\label{eq:adapter-objective}
\begin{aligned}
z_A &= \mathrm{LN}(E_A(x_A)),\qquad z_B = \mathrm{LN}(E_B(x_B)),\\
\mathcal{L} &=
\|D_A(z_A{+}\xi_A){-}x_A\|^2+\|D_B(z_B{+}\xi_B){-}x_B\|^2\\
&\quad+\lambda\|z_A{-}z_B\|^2 .
\end{aligned}
\end{equation}
The first and second terms of Eq.~\eqref{eq:adapter-objective} are the self-reconstruction errors in restoring the residuals of $A$ and
$B$ in the spaces of their own models. The third term is the latent alignment error, which pulls together $z_A$ and $z_B$ obtained at
the same token position. The larger $\lambda$ is, the larger the relative weight of the third term in the training loss $\mathcal L$.
Each squared norm is computed as a mean squared error over all tokens and dimensions.
$\mathrm{LN}$ is a non-affine layer normalization with no learnable scale or bias. $\xi_A,\xi_B\sim\mathcal N(0,\sigma^2 I)$
are independent Gaussian noise terms with mean $0$ added just before decoding. $\sigma>0$ is the hyperparameter that sets the standard
deviation of the noise, and $I$ is the $d_z{\times}d_z$ identity matrix.
In training the adapter we fix $\lambda{=}1$ and $\sigma{=}0.4$. Setting $\lambda{=}1$ gives the same coefficient $1$ to the three
mean-squared-error terms and does not mean that the terms take equal values during training. We did not tune these values per layer,
direction, or task, nor run ablation experiments that vary them.

The loss contains no term that predicts the residual of $B$ directly from that of $A$, nor one that predicts the residual of $A$ from that of $B$. In a cross-family
read-out, $z_A$ obtained with the encoder of $A$ is fed to the decoder of $B$ to give $D_B(z_A)$, and the read-out from $B$ to $A$ uses
$D_A(z_B)$. For these maps to work, the third term must bring $z_A$ and $z_B$ into the same coordinate system. Because the method uses
paired residuals, we do not call it unsupervised learning, even though no cross-family target is given directly.

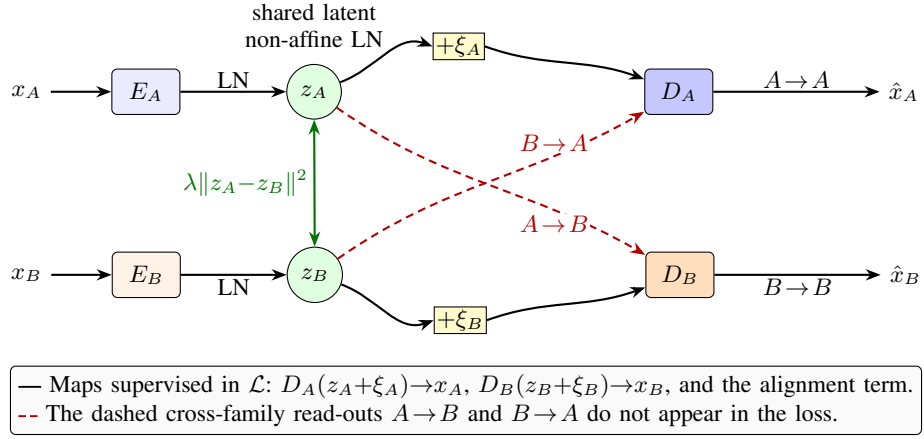
\begin{figure*}[!t]\centering
\begin{tikzpicture}[
  font=\small, >={Stealth[length=2mm]},
  box/.style={draw,rounded corners=2pt,minimum height=6mm,minimum width=9mm,align=center},
  encA/.style={box,fill=blue!8},   decA/.style={box,fill=blue!22},
  encB/.style={box,fill=orange!10}, decB/.style={box,fill=orange!26},
  lat/.style={draw,circle,minimum size=7mm,fill=green!12,align=center},
  sup/.style={->,thick}, cross/.style={->,thick,densely dashed,red!70!black}]
\node (xA) {$x_A$};
\node[encA,right=8mm of xA] (EA) {$E_A$};
\node[lat,right=14mm of EA] (zA) {$z_A$};
\node[decA,right=40mm of zA] (DA) {$D_A$};
\node[right=22mm of DA] (xAo) {$\hat x_A$};
\node[below=20mm of xA] (xB) {$x_B$};
\node[encB,right=8mm of xB] (EB) {$E_B$};
\node[lat,right=14mm of EB] (zB) {$z_B$};
\node[decB,right=40mm of zB] (DB) {$D_B$};
\node[right=22mm of DB] (xBo) {$\hat x_B$};
\node[draw,fill=yellow!25,inner sep=1pt,right=12mm of zA,yshift=6mm] (nA) {$+\xi_A$};
\node[draw,fill=yellow!25,inner sep=1pt,right=12mm of zB,yshift=-6mm] (nB) {$+\xi_B$};
\draw[sup] (xA)--(EA); \draw[sup] (EA)--node[above,inner sep=1pt]{LN} (zA);
\draw[sup] (xB)--(EB); \draw[sup] (EB)--node[below,inner sep=1pt]{LN} (zB);
\draw[sup] (zA) to[out=20,in=160] (nA.west); \draw[sup] (nA.east) to[out=-20,in=160] (DA);
\draw[sup] (zB) to[out=-20,in=200] (nB.west); \draw[sup] (nB.east) to[out=20,in=200] (DB);
\draw[sup] (DA)--node[above,inner sep=1pt]{$A\!\to\!A$}(xAo);
\draw[sup] (DB)--node[below,inner sep=1pt]{$B\!\to\!B$}(xBo);
\draw[<->,thick,green!45!black] (zA)--node[left,inner sep=2pt]{$\lambda\lVert z_A{-}z_B\rVert^2$} (zB);
\draw[cross] (zA) to[out=-35,in=150]
  node[pos=0.72,below=1pt,fill=white,inner sep=0.5pt] {$A\!\to\!B$} (DB);
\draw[cross] (zB) to[out=35,in=210]
  node[pos=0.72,above=1pt,fill=white,inner sep=0.5pt] {$B\!\to\!A$} (DA);
\node[align=center,above=1mm of zA] {shared latent\\non-affine LN};
\node[draw,rounded corners=2pt,fill=gray!4,inner sep=3pt,align=left,
      below=6mm of xB,anchor=north] at ($(xB)!0.5!(xBo)+(0,-6mm)$) {%
\textcolor{black}{\rule[0.4ex]{3mm}{0.7pt}}~Maps supervised in $\mathcal L$: $D_A(z_A{+}\xi_A){\to}x_A$, $D_B(z_B{+}\xi_B){\to}x_B$, and the alignment term.\\[1pt]
\textcolor{red!70!black}{\rule[0.4ex]{1mm}{0.7pt}\,\rule[0.4ex]{1mm}{0.7pt}}~The dashed cross-family read-outs $A\!\to\!B$ and $B\!\to\!A$ do not appear in the loss.};
\end{tikzpicture}
\caption{Shared-latent adapter. Solid lines are the self-reconstructions and the alignment that the loss includes, and dashed lines are the cross-family read-outs that the loss does not include.}
\label{fig:adapter}
\end{figure*}

Each encoder and decoder consists of an input linear projection, one residual block, and an output linear projection. The residual
block normalizes its input, applies a fully connected layer $\mathrm{fc}_1$, the activation function GELU (Gaussian Error Linear Unit),
and a fully connected layer $\mathrm{fc}_2$ in turn, and adds the result to its input, which is a pre-norm configuration.
Both the latent width and the output width of $\mathrm{fc}_1$ are $4096$. Because the weights of $\mathrm{fc}_2$ are zero-initialized,
the residual block starts as the identity map, and $E$ and $D$ begin training as linear maps. A non-affine LN lies between them,
however, so a cross-family read-out as a whole is not linear even at initialization.
Zero initialization is a property at the start of training and does not mean that the trained map is linear. Compared with the $14.25$B
parameters of the two parent models combined, the adapter has $268.5$M parameters, a ratio of $1.9\%$.

The noise $\xi$ is introduced to keep the latent from being dominated by a constant component that does not depend on the token.
We call this component the common mode.
The alignment term also becomes small when $z_A$ and $z_B$ consist only of this constant component. Adding noise
just before decoding, however, makes it hard to recover the original residuals from a latent with little per-token variation and thus
raises the self-reconstruction error. The noise is a design that relies on this property to make the latent keep its per-token
variation.

\subsection{Configurations and Chimera Models}\label{sec:config}
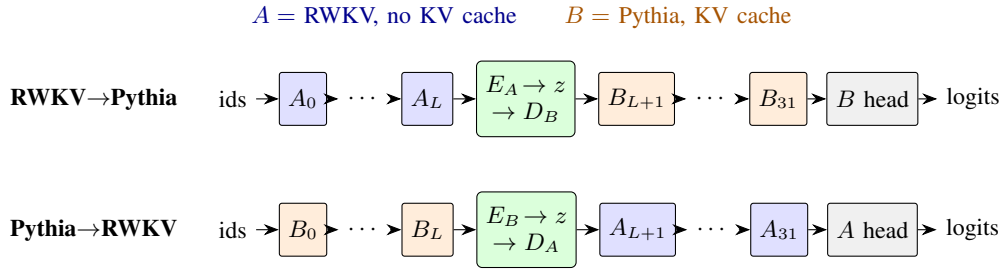
\begin{figure*}[!t]\centering
\begin{tikzpicture}[
  font=\small, >={Stealth[length=2mm]},
  a/.style={draw,rounded corners=1pt,minimum height=7mm,minimum width=6mm,fill=blue!12},
  b/.style={draw,rounded corners=1pt,minimum height=7mm,minimum width=6mm,fill=orange!14},
  hd/.style={draw,rounded corners=1pt,minimum height=7mm,minimum width=8mm,fill=gray!12},
  tr/.style={draw,rounded corners=2pt,minimum height=10mm,minimum width=13mm,fill=green!14,align=center}]
\node[font=\small\bfseries] (labAB) {\RtoP{}};
\node[right=3mm of labAB] (idsAB) {ids};
\node[a,right=3mm of idsAB] (a0) {$A_0$};
\node[right=1.5mm of a0] (ad) {$\cdots$};
\node[a,right=1.5mm of ad] (aL) {$A_L$};
\node[tr,right=3mm of aL] (tab) {$E_A\!\to z$\\$\to D_B$};
\node[b,right=3mm of tab] (bN) {$B_{L+1}$};
\node[right=1.5mm of bN] (bd) {$\cdots$};
\node[b,right=1.5mm of bd] (b31) {$B_{31}$};
\node[hd,right=2.5mm of b31] (hb) {$B$ head};
\node[right=2.5mm of hb] (outAB) {logits};
\draw[->] (idsAB)--(a0);
\foreach \x/\y in {a0/ad,ad/aL,aL/tab,tab/bN,bN/bd,bd/b31,b31/hb,hb/outAB} \draw[->] (\x)--(\y);
\node[font=\small\bfseries,below=12mm of labAB] (labBA) {\PtoR{}};
\node[right=3mm of labBA] (idsBA) {ids};
\node[b,right=3mm of idsBA] (bb0) {$B_0$};
\node[right=1.5mm of bb0] (bbd) {$\cdots$};
\node[b,right=1.5mm of bbd] (bbL) {$B_L$};
\node[tr,right=3mm of bbL] (tba) {$E_B\!\to z$\\$\to D_A$};
\node[a,right=3mm of tba] (aaN) {$A_{L+1}$};
\node[right=1.5mm of aaN] (aad) {$\cdots$};
\node[a,right=1.5mm of aad] (aa31) {$A_{31}$};
\node[hd,right=2.5mm of aa31] (ha) {$A$ head};
\node[right=2.5mm of ha] (outBA) {logits};
\draw[->] (idsBA)--(bb0);
\foreach \x/\y in {bb0/bbd,bbd/bbL,bbL/tba,tba/aaN,aaN/aad,aad/aa31,aa31/ha,ha/outBA} \draw[->] (\x)--(\y);
\node[above=3mm of tab,align=center] {
  \textcolor{blue!55!black}{$A=$ RWKV, no KV cache}\qquad
  \textcolor{orange!65!black}{$B=$ Pythia, KV cache}
};
\end{tikzpicture}
\caption{Configurations of the \RtoP{} and \PtoR{} chimera models.}
\label{fig:chimera}
\end{figure*}

We call the layer at which the residual of the front model is converted by the adapter and passed to the back model the switch layer, the model that runs up to the switch
layer the front model, and the model that runs after the switch layer the back model. A combination of a front and a back model is
called a configuration, including the case in which the same model is used for both; a model with a specified switch layer and a
parent model alone are also called configurations.
The front-back combinations compared in this study are of four kinds: \RWKVself{}, \RtoP{}, \Pythiaself{}, and \PtoR{}.
\RtoP{} and \PtoR{} are cross-family chimera models, and \RWKVself{} and \Pythiaself{} are self-reconstruction controls that pass
through the same adapter at the same switch layer. The switch layer $L$ is an index counted from $0$ and denotes the last block that
the front parent model runs. Because block $0$ is counted, the front model runs $L{+}1$ blocks.
In the notation of \S\ref{sec:notation}, the conversion at the switch layer is given by the following equation, where a tilde marks
a residual converted into the coordinate system of the other model.

\begin{equation}\label{eq:boundary}
\begin{aligned}
\widetilde h_B^L[t]
 &= \mathrm{unstd}_B^L\!\left(D_B\!\left(\mathrm{LN}(E_A(x_A^L[t]))\right)\right),\\
\widetilde h_A^L[t]
 &= \mathrm{unstd}_A^L\!\left(D_A\!\left(\mathrm{LN}(E_B(x_B^L[t]))\right)\right).
\end{aligned}
\end{equation}
Block $L{+}1$ of $B$ receives $\widetilde h_B^L[t]$, and block $L{+}1$ of $A$ receives $\widetilde h_A^L[t]$.
Equation~\eqref{eq:boundary} adds no training noise. Execution has three steps: running blocks $0$ to $L$ of the front parent model,
converting the residual once, and running blocks $L{+}1$ to $31$ and the language-model head of the back parent model, as shown in
\figref{fig:chimera}. The conversion is applied only once, at the switch layer, with no round trip at each block.
Every chimera model therefore runs $32$ blocks in total, $L{+}1$ in the front model and $31{-}L$ in the back model. The error in the
residual introduced by the conversion at the switch layer is called the conversion error.
Which model serves as the front model is called the connection direction. In \PtoRL{4}, for example, the $5$ blocks $0$ to $4$ of
Pythia run first, followed by the $27$ blocks $5$ to $31$ of RWKV.

A configuration is denoted by the names of its front and back models, as in \RtoP{}, and the switch layer is added as in \RtoPL{4}.
Read-outs, in contrast, are denoted by symbols such as $A\!\to\!B$. The read-out used at the switch layer of \RtoP{} is $A\!\to\!B$,
that used by \PtoR{} is $B\!\to\!A$, and those used by \RWKVself{} and \Pythiaself{} are $A\!\to\!A$ and $B\!\to\!B$, respectively.
In tables, RWKV is abbreviated as R and Pythia as P, and \RtoP{} and \RWKVself{} are written, for example, as R$\to$P
and R self.

\subsection{Evaluation Metrics}\label{sec:common-mode}
How well a residual is recovered is measured by the coefficient of determination $\Rsq=1-\mathrm{SSE}/\mathrm{SST}$. $\mathrm{SSE}$ is
the sum of squared errors between the predicted and true residuals, and
$\mathrm{SST}$ is the total sum of squares of the true residuals after subtracting the per-dimension token mean. A value of $1$ means
perfect recovery, $0$ means the same performance as always predicting the per-dimension mean, and a negative value means worse than
this mean predictor. We measure four read-outs: the cross-family $A\!\to\!B$ and $B\!\to\!A$ and the self-reconstructions
$A\!\to\!A$ and $B\!\to\!B$. The latter two are controls that show the information lost by encoding and decoding.
Below, the $\Rsq$ of a read-out is called its read-out accuracy.

$\Rsq$ is computed per layer over the evaluation tokens and all $4096$ dimensions together. Because $\mathrm{SST}$ is the total
variation after removing the per-dimension token mean, dimensions with larger variance contribute more. The alignment of the latent
spaces is measured by the centered correlation $\rho_{\mathrm{ctr}}$, the Pearson correlation coefficient of $z_A,z_B$ flattened into
one dimension after subtracting the token mean from each dimension. Values close to $1$ mean that the two latents vary in the same
direction, and $0$ means no linear correlation. We further define the varying fraction, the fraction of the per-token varying component of the latent, as $f=\lVert z-\overline z\rVert^2/\lVert z\rVert^2$, where
$\overline z$ is the per-dimension token mean within a layer and $f$ is computed separately for $A$ and $B$ and averaged. Each metric
is computed per layer, and we report the simple mean over the $32$ layers.

There are two sets of evaluation data, both taken from positions that do not overlap the training sequences. The main values in the
text are computed on sequences of $112$ tokens, $1{,}000$ sequences or $112{,}000$ tokens in total.
These data are also used to adjust the learning rate and to select the checkpoint (\S\ref{sec:train}). The per-layer analysis and the
shuffle control, which re-pairs the tokens at random, use $400$ sequences of the same length, $44{,}800$ tokens in total.
Because the evaluation data differ, the $32$-layer means of the per-layer analysis differ slightly from the main values; both are
given in \S\ref{sec:space}.
The affine control for accuracy evaluation in \S\ref{sec:linear} is fitted on $40{,}096$ rows of standardized residuals taken from the
part of Alpaca not used to train the adapter. Each row is the residual vector at one token position. These $40{,}096$ rows and the sequences used for the linearity evaluation and the
intervention in \S\ref{sec:linear} are leading parts of the main evaluation data, taken from the same positions by the same procedure, and are contained in it.
The fitted $R^2$ of the all-layer correspondence expresses how well the representation of one layer can be predicted linearly from
that of another. A linear map is fitted by ridge regression, that is, least squares with L2 regularization, to the $22{,}400$ rows shared by each
pair of layers, which come from $200$ sequences of $112$ tokens taken from the same positions as the main evaluation data. The same
samples are used for fitting and evaluation, so for a $4096$-dimensional linear map the value is easily inflated.

The language ability of the chimera models is evaluated in three ways. First, English prompts are given without modification, and
generation examples are obtained by deterministic argmax decoding at temperature $0$, which selects the most probable token at each
step. Generation is limited to $40$ new tokens and stops early only when the model outputs EOS (end-of-sequence), the special token
that marks the end of a sequence. Second, the log-likelihood of each option of a multiple-choice question is computed, the most likely
option is taken as the answer, and accuracy is measured. The question sets are ARC-Easy~\cite{clark2018thinksolvedquestionanswering},
the subset of the ARC (AI2 Reasoning Challenge) dataset of science exam questions for elementary and middle school students that
excludes the hard questions both a retrieval-based and a word co-occurrence baseline answered incorrectly, and
SciQ~\cite{welbl-etal-2017-crowdsourcing}. Below, ARC-Easy is abbreviated as ARC. Both mostly
have 4 options, and random choice gives an accuracy of about $25\%$.

We use all $2376$ test questions of ARC and all $1000$ test questions of SciQ.
SciQ is evaluated in closed-book form, without its support passages, so our SciQ values cannot be compared directly with published
values obtained under the standard protocol that provides those passages.
Two scores are computed from the log-likelihoods of the option tokens conditioned on the question, that is, of the tokens after the
prompt: their sum and their average.
Options differ in length and the sum tends to be smaller for longer options, so the text mainly uses the latter, the length-normalized
score. Differences between configurations are computed from paired data on the same questions, so they are evaluated with McNemar's exact test
for paired binary data and with a paired bootstrap that estimates the variability of the difference by resampling pairs of the same
questions. The text and tables give McNemar $p$-values, and bootstrap confidence intervals of the differences are included in the
released data. Accuracy differences are expressed in percentage points, called points below.

Third, the next-token cross-entropy is measured on Alpaca and on WikiText-103~\cite{merity2017pointer}, a
corpus built from good-quality Wikipedia articles and called WikiText below, to compare language-modeling performance inside and outside the domain of text used to train the
adapter, called the training domain below.

\subsection{Training Setup}\label{sec:train}
The adapter is trained on the residuals of $2{,}075$ sequences of $112$ tokens taken from the beginning of the Alpaca training split.
The distinct token positions number about $232$k, and the residual pairs number $7.44$M over the $32$ layers. This set of pairs was
iterated about $228$ times.
For optimization we used AdamW, which decouples weight decay from Adam (Adaptive Moment Estimation), with an effective batch size of
$4096$, an initial learning rate of $10^{-3}$, and a minimum learning rate of $10^{-5}$.
At each iteration one of the $32$ layers is chosen uniformly at random, and an effective batch is drawn from the residual pairs of the
chosen layer. Gradients are clipped so that their norm does not exceed $1.0$.
Every $1{,}000$ iterations, $A\!\to\!B$ was measured on the evaluation data (\S\ref{sec:common-mode}), and a reference value was updated only by
an evaluation that exceeded it by more than a relative $10^{-3}$; the learning rate was multiplied by $0.5$ each time $4$ consecutive
evaluations brought no update. We ran $416{,}000$ iterations with random seed $0$ and
used the checkpoint at iteration $415{,}000$, where $A\!\to\!B$ on the evaluation data was best, for all subsequent evaluations.
The checkpoint was not reselected based on $B\!\to\!A$ or on multiple-choice results. Details of the implementation and the released
artifacts are given in Appendix~\ref{sec:repro}.

\section{Results}\label{sec:results}
We first examine whether the representations of corresponding layers align, and measure how well each model's residuals can be
recovered from the resulting representation. We then analyze the nonlinearity of the adapter, and finally evaluate the accuracy,
inference memory, performance outside the training domain, and generation examples of the composed chimera models.

\subsection{Shared Representation Between Corresponding Layers}\label{sec:space}
The adapter trained on the residuals of both models shows a centered correlation of $\rho_{\mathrm{ctr}}\!=\!0.901$ on text not used
for training. In addition to this value, we take as evidence that a shared representation exists the fact that the cross-family
read-out accuracy is $0.559$ for $A\!\to\!B$ and $0.590$ for $B\!\to\!A$ and that both fall sharply under the shuffle control
described next. The self-reconstruction read-out accuracies, $A\!\to\!A\ 0.695$ and $B\!\to\!B\ 0.711$, indicate how much information
is lost by encoding and decoding.

This shared representation is based on per-token correspondence rather than on overlapping distributions. When the row order of $A$ is
kept and the residual rows of $B$ are re-paired under a single random permutation, on the evaluation data for the per-layer analysis
$\rho_{\mathrm{ctr}}$ falls as $0.901\!\to\!0.000$, and the cross-family read-out accuracy falls as $0.561\!\to\!-0.592$ for $A\!\to\!B$
and as $0.592\!\to\!-0.609$ for $B\!\to\!A$.
The self-reconstruction read-out accuracies do not change, by construction. Neither cross-family read-out therefore succeeds without the
correct token correspondence.

\begin{figure*}[!t]\centering
\includegraphics[width=\textwidth]{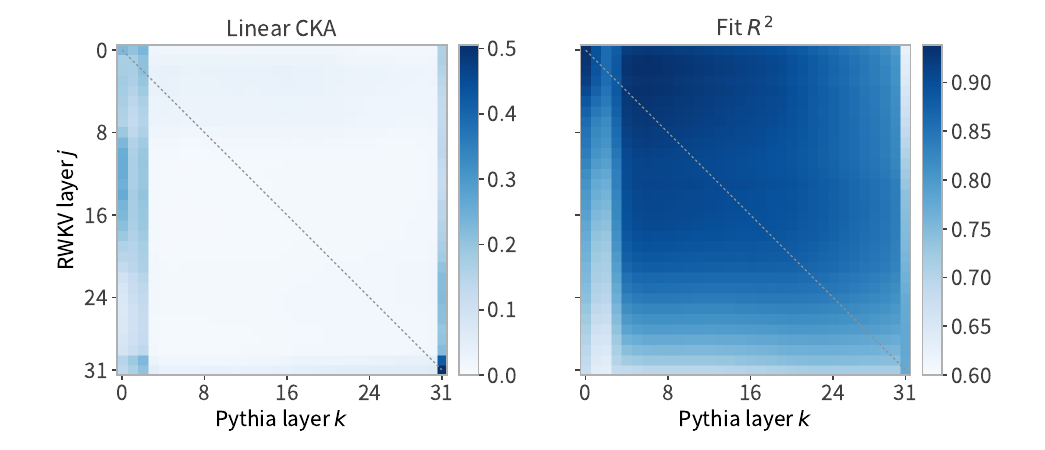}
\caption{Correspondence over all $32{\times}32$ layer pairs of the unstandardized residuals $h$ taken without the adapter. Left: linear CKA. Right: fitted $R^2$ of per-layer
linear maps. Dashed lines mark same-index layers, and the two panels use different color scales.}
\label{fig:cka}
\end{figure*}

Same-index layer pairs, however, are not intrinsically special. As \figref{fig:cka} shows, comparing the unstandardized residuals $h$ directly, without
the adapter, over all $32{\times}32$ layer pairs gives a small linear CKA overall. The mean of the diagonal elements, which are the
pairs of layers with the same index, is $0.05$, and the mean over the other pairs is $0.03$.
Layer $j$ of $A$ is most similar to layer $j$ of $B$ in only $2$ of $32$ rows. For $14$ of the $32$ layers of $A$ the most similar layer of $B$ is
layer $0$, and for $12$ it is the last layer. Same-index layers are not special in linear predictability either.
Fitting a linear map to every layer pair gives fitted $R^2$ values of $0.62$--$0.94$, high for almost every pair. These values are
measured on the rows used for fitting, however, and cannot be compared with values evaluated on rows not used for fitting, such as those of the direct affine maps
in \S\ref{sec:linear}.
The mean fitted $R^2$ is $0.87$ on the diagonal and $0.84$ off the diagonal, and
the same-index layer is best in only $3$ of $32$ rows. The value $\rho_{\mathrm{ctr}}\!=\!0.901$ is therefore an alignment obtained by
training on the chosen pairs of same-index layers, not a correspondence inherent in layers that share an index.
In other words, this study is a case study that connects same-index layer pairs in a setting where any layer pair can be predicted by
linear regression to some degree.

The Pearson correlation coefficient of $z_A,z_B$ before centering, called the raw correlation below, is $\rho_{\mathrm{raw}}=0.965$,
which is higher than the centered correlation $0.901$.
The difference is explained by the common mode. With the varying fraction $f$ defined in \S\ref{sec:common-mode}, the following holds
under a layer normalization with no per-element coefficients when the varying fractions of $A$ and $B$ are equal and their common
modes are parallel.
\begin{equation}\label{eq:common-mode}
\rho_{\mathrm{raw}} = (1-f)+f\,\rho_{\mathrm{ctr}}.
\end{equation}
Here $f$ is the mean of the varying fractions of $A$ and $B$. At evaluations after training has progressed,
Equation~\eqref{eq:common-mode} holds to within about $3\!\times\!10^{-4}$. At iteration $0$, however, the error is $6.6\!\times\!10^{-2}$.
The checkpoint at iteration $415{,}000$ has $f=0.350$, so the token-independent common mode accounts for $65\%$ of the squared norm of the latent.
Equation~\eqref{eq:common-mode} reproduces the raw correlation exactly as $(1{-}0.350)+0.350\times0.901=0.965$, which shows that most of
the raw correlation is due to the common mode, which contains no information.
We therefore use the centered correlation as the measure of alignment.

Note that $f$ is a second moment, not an amount of information. $f\!\to\!0$ means that the latent has degenerated to a constant and
contains no per-token information, whereas a large $f$ only shows that the latent varies in direction. Whether the varying component
contains information is judged by the read-outs.
The read-out measure is a centered $\Rsq$, which no token-independent predictor can raise above $0$. Hence $B\!\to\!B=0.711$ cannot be
obtained from a constant prediction, which is consistent with the common mode accounting for $65\%$ of the squared norm of the latent.

\subsection{Self-Reconstruction Compared with Cross-Family Read-Outs}\label{sec:ladder}
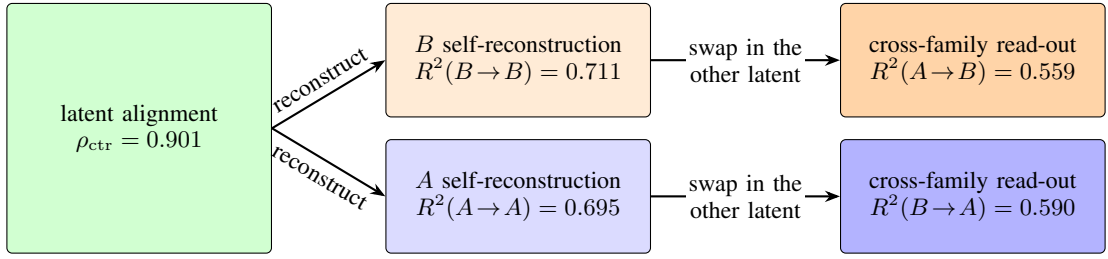
\begin{figure*}[!t]\centering
\begin{tikzpicture}[
  font=\small, >={Stealth[length=2mm]},
  diag/.style={draw,rounded corners=2pt,minimum width=35mm,minimum height=15mm,align=center}]
  \node[diag,fill=green!18,minimum height=33mm] (align) {latent alignment\\$\rho_{\mathrm{ctr}}=0.901$};
  \node[diag,fill=orange!16,right=15mm of align,yshift=9mm] (selfb)
    {$B$ self-reconstruction\\$\Rsq(B\!\to\!B)=0.711$};
  \node[diag,fill=orange!34,right=25mm of selfb] (crossab)
    {cross-family read-out\\$\Rsq(A\!\to\!B)=0.559$};
  \node[diag,fill=blue!14,right=15mm of align,yshift=-9mm] (selfa)
    {$A$ self-reconstruction\\$\Rsq(A\!\to\!A)=0.695$};
  \node[diag,fill=blue!30,right=25mm of selfa] (crossba)
    {cross-family read-out\\$\Rsq(B\!\to\!A)=0.590$};
  \draw[->,thick] (align.east)--node[above,sloped]{reconstruct} (selfb.west);
  \draw[->,thick] (align.east)--node[below,sloped]{reconstruct} (selfa.west);
  \draw[->,thick] (selfb)--node[midway,fill=white,inner sep=1pt,align=center]
    {swap in the\\other latent} (crossab);
  \draw[->,thick] (selfa)--node[midway,fill=white,inner sep=1pt,align=center]
    {swap in the\\other latent} (crossba);
\end{tikzpicture}
\caption{Relation between the shared latent space and the four read-outs. Read-outs of the same color share a decoder.}
\label{fig:ladder}
\end{figure*}

\figref{fig:ladder} shows the relation among the five measures described in \S\ref{sec:space}.
$\rho_{\mathrm{ctr}}$ is a correlation coefficient and the other four are coefficients of determination, so they cannot be compared on
the same scale. Among the cross-family read-out accuracies, $B\!\to\!A$, used by \PtoR{}, slightly exceeds $A\!\to\!B$, used by
\RtoP{}, so the RWKV residuals are easier to recover from the latent obtained with the Pythia encoder. Both read-outs fall below the
self-reconstruction that uses the same decoder and differs only in the input latent.
For the read-out accuracies, $A\!\to\!B\le B\!\to\!B$ and $B\!\to\!A\le A\!\to\!A$ hold in all $32$ layers.
This ordering is observed in experiments rather than guaranteed by theory; still, in both directions substantial error already
exists at
the self-reconstruction stage, and improving the cross-family part does not remove the self-reconstruction error.

At $d_z{=}d{=}4096$ the latent is as wide as the residual, and the non-affine LN removes only two degrees of freedom per token, the
mean and the norm. Latent width or rank is therefore unlikely to constrain reconstruction strongly. Even so, $B\!\to\!B$ is only $0.711$.
Whether this is an upper bound set by the noise $\sigma$ is unknown. We have also not isolated whether the error arises in the encoder,
the latent normalization, the decoder, or the training noise, and what limits the reconstruction accuracy remains unresolved.

\subsection{Nonlinearity of the Adapter}\label{sec:linear}
As described in \S\ref{sec:adapter}, $E$ and $D$ of the adapter begin training as linear maps.
This is a property before training, however, not after $415{,}000$ iterations. We therefore examine the linearity of the adapter at
this checkpoint with three affine maps that play different roles. Maps (i) and (ii) were measured for both \RtoP{} and \PtoR{} on
the same $5$ representative layers $\{4,10,16,22,28\}$ and the same fitting and evaluation rows. The $5$ representative layers were
taken every $6$ layers from layer $4$ to layer $28$. The rows are standardized residuals of
$900$ sequences of $112$ tokens taken from positions of Alpaca not used to train the adapter; for each layer the first $50{,}400$ rows
are used for fitting and the last $50{,}400$ rows for evaluation.

(i) The imitation affine map is the affine map that best approximates the output of the trained adapter, and it is used to measure how
linear the adapter is. The fraction of the adapter's output variance that it explains is
$86.6\%$ for \RtoP{} and $88.0\%$ for \PtoR{}. The remaining $13.4\%$ and $12.0\%$ therefore cannot be explained on the evaluation rows by the affine map fitted on the first half of
the rows, and
the trained map is nonlinear in both directions.

(ii) The direct affine map predicts, without the adapter, the standardized output residual from the standardized input residual for
each layer and direction; it is the best affine map fitted by ridge regression. The regularization coefficient is
$10^{-3}$ times $\mathrm{tr}(X^{\top}X)/(d{+}1)$ for the input matrix $X$ with an intercept column appended. This map measures how much
cross-family prediction a low-capacity map achieves. On the evaluation rows,
the best direct affine map gives $0.487$ for \RtoP{} and $0.535$ for \PtoR{}. On the same rows the adapter gives $0.558$ for \RtoP{}
and $0.576$ for \PtoR{}, so
the nonlinear map adds $+0.071$ and $+0.041$. These values were evaluated on the $5$ representative layers, and both the layers
and the rows differ from those of the $32$-layer means $0.559$ and $0.590$ in \S\ref{sec:space}. That even a low-capacity affine map predicts
about half of the variance of the other model's residual in both directions, $87\%$ and $93\%$ of the adapter's values, is consistent with the
conclusion that a shared representation exists.

(iii) The affine control for accuracy evaluation embeds an affine map like (ii) in the chimera model and serves as a control for
comparing multiple-choice accuracy with the adapter.
It was fitted for each switch layer on the rows described in \S\ref{sec:common-mode}, by ridge regression with an intercept
and the same regularization coefficient as (ii). As \tabref{tab:bench} shows,
replacing the adapter of \RtoP{} with this control lowers accuracy by $8$--$23$ points. For \PtoR{}, by contrast, the difference is
small and lies within $-3.4$--$+2.0$ points over the $8$ conditions.
At $L{=}4$ on ARC the adapter is $2.0$ points higher ($p{=}0.005$), whereas at $L{=}8$ on SciQ the affine control is $3.4$ points
higher ($p{=}0.003$). The two maps differ in training objective, data, and
regularization, so we do not treat this difference between directions as a causal effect of nonlinearity.

\begin{table}[!t]\centering\small
\caption{Length-normalized accuracy (\%) of the adapter (Ad) and of the affine control for accuracy evaluation (Af). R$\to$P: RWKV$\to$Pythia; P$\to$R: Pythia$\to$RWKV.}\label{tab:bench}
\setlength{\tabcolsep}{2.4pt}
\begin{tabular}{rcccccccc}\toprule
& \multicolumn{4}{c}{ARC} & \multicolumn{4}{c}{SciQ}\\
\cmidrule(lr){2-5}\cmidrule(lr){6-9}
& \multicolumn{2}{c}{R$\to$P} & \multicolumn{2}{c}{P$\to$R}
& \multicolumn{2}{c}{R$\to$P} & \multicolumn{2}{c}{P$\to$R}\\
$L$ & Ad & Af & Ad & Af & Ad & Af & Ad & Af\\\midrule
4  & 59.6 & 41.9 & 62.2 & 60.2 & 70.9 & 47.5 & 69.2 & 69.6\\
8  & 59.0 & 43.1 & 57.3 & 56.0 & 67.7 & 49.1 & 64.6 & 68.0\\
16 & 48.6 & 34.8 & 57.5 & 57.5 & 54.5 & 35.9 & 68.0 & 69.6\\
24 & 47.0 & 39.4 & 59.6 & 59.2 & 51.1 & 42.7 & 69.9 & 68.2\\\bottomrule
\end{tabular}
\end{table}

Next, we intervene to disable the nonlinearity of the adapter. To suppress the confounds above, in each residual block of the same trained
adapter we scale by a coefficient $\alpha$ only the output of the branch that normalizes the input and applies $\mathrm{fc}_1$, GELU,
and $\mathrm{fc}_2$. This branch is the only learned nonlinearity in the adapter, and we call its output the nonlinear part. $\alpha{=}1$ is the original map, and
$\alpha{=}0$ is the same map with the nonlinear part removed.
Sweeping $\alpha=0,.25,.5,.75,1$ for the mean cross-family read-out accuracy of the $5$ representative layers on the evaluation data for
the per-layer analysis (\S\ref{sec:common-mode}) gives
$-0.41,-0.75,0.35,0.52,0.56$ for \RtoP{} and $-0.30,-0.66,0.36,0.53,0.58$ for \PtoR{}.
Removing the nonlinear part lowers the read-out accuracy to negative values in both directions, and the dependence on $\alpha$ is
non-monotonic. Even at $\alpha{=}0$ the non-affine LN remains, so the map does not become entirely linear. The input and output linear projections of the encoder and the decoder were
trained jointly with the nonlinear part, and in this trained adapter the nonlinear part is indispensable to both the \RtoP{} and
\PtoR{} read-outs.

\begin{table}[!t]\centering\small
\caption{Change in length-normalized accuracy (\%) when the nonlinear part is disabled, $\alpha:1\!\to\!0$. R$\to$P: RWKV$\to$Pythia; P$\to$R: Pythia$\to$RWKV.}\label{tab:mlp-intervention}
\setlength{\tabcolsep}{3pt}
\begin{tabular}{rcccc}\toprule
& \multicolumn{2}{c}{ARC} & \multicolumn{2}{c}{SciQ}\\
\cmidrule(lr){2-3}\cmidrule(lr){4-5}
$L$ & R$\to$P & P$\to$R & R$\to$P & P$\to$R\\\midrule
4  & $59.6\!\to\!31.9$ & $62.2\!\to\!31.8$ & $70.9\!\to\!31.3$ & $69.2\!\to\!32.6$\\
8  & $59.0\!\to\!29.9$ & $57.3\!\to\!35.5$ & $67.7\!\to\!31.4$ & $64.6\!\to\!37.0$\\
16 & $48.6\!\to\!28.5$ & $57.5\!\to\!43.7$ & $54.5\!\to\!31.7$ & $68.0\!\to\!47.3$\\
24 & $47.0\!\to\!34.8$ & $59.6\!\to\!48.4$ & $51.1\!\to\!40.0$ & $69.9\!\to\!49.1$\\\bottomrule
\end{tabular}
\end{table}

The effect of the nonlinear part in \tabref{tab:mlp-intervention} is $+11.1$--$+39.6$ points for \RtoP{} and
$+11.2$--$+36.6$ points for \PtoR{}, and McNemar's exact test on the same questions gives
$p<2{\times}10^{-9}$ in all $16$ comparisons. Removing the nonlinear part from this trained adapter therefore lowers the task accuracy
as well as the read-out accuracy for both \RtoP{} and \PtoR{}. This is, however, a property of removing the nonlinear part from a
jointly trained map, and it does not show that nonlinearity is generally necessary for connecting families.
We did not compare against a linear adapter trained for the same objective, and the accuracy difference between the affine control and the
adapter is small for \PtoR{} (\tabref{tab:bench}).

\subsection{Accuracy Compared with the Parent Models}\label{sec:goal}
This subsection compares the chimera models with the two frozen parent models to examine how much accuracy is retained relative to
the stronger parent and whether the weaker parent is exceeded. The evaluation method is as described in \S\ref{sec:common-mode}.

\begin{table}[!t]\centering\small
\caption{Length-normalized accuracy (\%) of the four configurations. The maximum of each row and task is in bold. R self: RWKV self-reconstruction control; P self: Pythia self-reconstruction control; R$\to$P: RWKV$\to$Pythia; P$\to$R: Pythia$\to$RWKV.}\label{tab:direction}
\setlength{\tabcolsep}{2.4pt}
\begin{tabular}{rcccccccc}\toprule
& \multicolumn{2}{c}{R self} & \multicolumn{2}{c}{R$\to$P}
& \multicolumn{2}{c}{P self} & \multicolumn{2}{c}{P$\to$R}\\
\cmidrule(lr){2-3}\cmidrule(lr){4-5}\cmidrule(lr){6-7}\cmidrule(lr){8-9}
$L$ & ARC & SciQ & ARC & SciQ & ARC & SciQ & ARC & SciQ\\\midrule
4  & \textbf{62.8} & 68.5 & 59.6 & 70.9 & 62.2 & \textbf{74.8} & 62.2 & 69.2\\
8  & 59.8 & 66.7 & 59.0 & 67.7 & \textbf{62.7} & \textbf{74.8} & 57.3 & 64.6\\
16 & 54.8 & 59.8 & 48.6 & 54.5 & 57.4 & \textbf{71.1} & \textbf{57.5} & 68.0\\
24 & 52.2 & 57.9 & 47.0 & 51.1 & 59.4 & \textbf{70.1} & \textbf{59.6} & 69.9\\\bottomrule
\end{tabular}
\end{table}

\begin{table}[!t]\centering\small
\caption{Accuracy differences (points) between the main configurations and McNemar exact-test $p$-values in parentheses. R$\to$P: RWKV$\to$Pythia; P$\to$R: Pythia$\to$RWKV.}\label{tab:paired}
\setlength{\tabcolsep}{1.6pt}
\begin{tabular}{rcccccc}\toprule
& \multicolumn{3}{c}{ARC} & \multicolumn{3}{c}{SciQ}\\
\cmidrule(lr){2-4}\cmidrule(lr){5-7}
& \multicolumn{2}{c}{vs RWKV} & Dir. & \multicolumn{2}{c}{vs RWKV} & Dir.\\
\cmidrule(lr){2-3}\cmidrule(lr){5-6}
$L$ & R$\to$P & P$\to$R & \shortstack{P$\to$R\\$-$R$\to$P} & R$\to$P & P$\to$R & \shortstack{P$\to$R\\$-$R$\to$P}\\\midrule
4 & $-2.2$ & $+0.4$ & $+2.6$ & $+3.9$ & $+2.2$ & $-1.7$\\
   & ($0.042$) & ($0.66$) & ($0.015$) & ($0.018$) & ($0.14$) & ($0.28$)\\[2pt]
8 & $-2.8$ & $-4.5$ & $-1.7$ & $+0.7$ & $-2.4$ & $-3.1$\\
   & ($0.007$) & ($<10^{-4}$) & ($0.11$) & ($0.69$) & ($0.081$) & ($0.052$)\\[2pt]
16 & $-13.2$ & $-4.3$ & $+8.9$ & $-12.5$ & $+1.0$ & $+13.5$\\
   & ($<10^{-4}$) & ($<10^{-4}$) & ($<10^{-4}$) & ($<10^{-4}$) & ($0.50$) & ($<10^{-4}$)\\[2pt]
24 & $-14.8$ & $-2.2$ & $+12.6$ & $-15.9$ & $+2.9$ & $+18.8$\\
   & ($<10^{-4}$) & ($0.043$) & ($<10^{-4}$) & ($<10^{-4}$) & ($0.082$) & ($<10^{-4}$)\\[2pt]
\bottomrule
\end{tabular}
\end{table}

\tabref{tab:direction} gives the accuracies of the four configurations, and \tabref{tab:paired} gives paired tests for the main
comparisons. In \tabref{tab:paired}, ``vs RWKV'' is the accuracy difference between each configuration and RWKV alone, and ``Dir.'' is
the difference between \PtoR{} and \RtoP{} at the same switch layer; both are based on the length-normalized score. The comparisons with Pythia
alone are omitted from the table. The main results are the following five points.

First, even though the front and back models come from different families, both \RtoP{} and \PtoR{} answer the questions, and every
configuration exceeds the accuracy of random choice, about $25\%$, by a wide margin.
Second, the chimera models are significantly less accurate than the stronger parent Pythia, whose accuracy is ARC $65.3\%$ and SciQ $81.0\%$
(\tabref{tab:kv}), in all $16$ conditions over both directions, all switch layers, and both tasks ($p\le0.003$).
Third, among the chimera models, the only one that significantly exceeds RWKV, whose accuracy is ARC $61.8\%$ and SciQ $67.0\%$, is \RtoPL{4} on SciQ,
and \PtoR{} does not significantly exceed RWKV at any switch layer.
The advantage of \RtoPL{4} appears only with the length-normalized score; with the summed likelihood score, the differences from RWKV alone are $-2.7$ points
on SciQ for \RtoPL{4}, and $-4.6$ points on ARC and $-5.7$ points on SciQ for \PtoRL{4}.
Fourth, the effect of the switch depth depends on the direction. \RtoP{} loses accuracy monotonically as the switch layer deepens,
whereas \PtoR{} is non-monotonic and substantially exceeds \RtoP{} at the same switch layer for $L{=}16,24$. On the evaluation data for the per-layer
analysis, the cross-family read-out accuracies at the same four layers are
$0.570/0.555/0.575/0.567$ for \RtoP{} and $0.552/0.517/0.604/0.608$ for \PtoR{}, so the accuracy ranking cannot be predicted from
read-out accuracy alone.
Fifth, the loss from encoding and decoding also appears in the accuracy of the self-reconstruction controls: the difference from
Pythia alone is significant for \Pythiaself{} at every switch layer on both tasks, and the difference from RWKV alone is significant for
\RWKVself{} on both tasks at $L{=}16,24$. At $L{=}8$ it is significant only on ARC ($p{=}0.006$). The differences from the
self-reconstruction controls, however, are not due to conversion error alone. With Pythia as the back model, accuracy decreases in
the order of Pythia alone, \Pythiaself{}, and \RtoP{}, and the difference between \Pythiaself{} and \RtoP{} is $3$--$12$ points on ARC
and $4$--$19$ points on SciQ. With RWKV as the back model, in contrast, \PtoR{} exceeds \RWKVself{} at deep switch layers, by
$+2.7/+7.4$ points on ARC and $+8.2/+12.0$ points on SciQ at $L{=}16/24$.
A difference from a cross-family configuration also contains the difference in ability between the front parent models, so it cannot
be read as the loss caused by conversion.

\subsection{KV Cache Reduction}\label{sec:kv}
For each layer and token, Pythia's KV cache holds two $4096$-dimensional vectors, the key and the value, in FP16, totaling $16$ KiB,
or $512$ KiB over $32$ layers.
RWKV has no KV cache but instead keeps a recurrent state of $64$ KiB per layer that does not depend on the context length. This value
is computed from the reference implementation, which keeps $2$ of its $5$ state vectors in FP16 and $3$ in 32-bit floating point (FP32). The KV cache sizes in
\tabref{tab:kv} exclude the recurrent state.
\RtoP{} keeps $31{-}L$ Transformer layers, and \PtoR{} keeps $L{+}1$. The reduction applies to the KV cache, not to the resident weights.
The adapter itself occupies $0.50$ GiB in FP16 and about $0.25$ GiB even when only one direction is kept in memory, which is comparable
to the reduced KV cache.
Even when the unused blocks are physically removed, the resident weights of a chimera model, including the adapter, are $13.49$--$14.55$ GiB, about the same as the $13.77$ GiB of RWKV
alone and the $12.77$ GiB of Pythia alone. The output of the implementation with the blocks removed matched that of the implementation without removal.

\begin{table}[!t]\centering\small
\caption{Accuracy and inference memory. KV@4k: KV cache size at context length $4096$. P alone: Pythia; R alone: RWKV; R$\to$P: RWKV$\to$Pythia; P$\to$R: Pythia$\to$RWKV.}\label{tab:kv}
\setlength{\tabcolsep}{2.4pt}
\begin{tabular}{lrrrrrr}\toprule
Config. & $L$ & \shortstack{Trans-\\former\\layers} & \shortstack{KV/token\\(KiB)} & \shortstack{KV@4k\\(GiB)} & \shortstack{Reduc.\\(\%)} & \shortstack{ARC/\\SciQ}\\\midrule
P alone & -- & 32 & 512 & 2.00 & 0 & 65.3/81.0\\
R$\to$P & 4  & 27 & 432 & 1.69 & 15.6 & 59.6/70.9\\
R$\to$P & 8  & 23 & 368 & 1.44 & 28.1 & 59.0/67.7\\
R$\to$P & 16 & 15 & 240 & 0.94 & 53.1 & 48.6/54.5\\
R$\to$P & 24 &  7 & 112 & 0.44 & 78.1 & 47.0/51.1\\
P$\to$R & 24 & 25 & 400 & 1.56 & 21.9 & 59.6/69.9\\
P$\to$R & 16 & 17 & 272 & 1.06 & 46.9 & 57.5/68.0\\
P$\to$R & 8  &  9 & 144 & 0.56 & 71.9 & 57.3/64.6\\
P$\to$R & 4 & 5 & 80 & 0.31 & 84.4 & 62.2/69.2\\
R alone & -- & 0 & 0 & 0 & 100 & 61.8/67.0\\\bottomrule
\end{tabular}
\end{table}

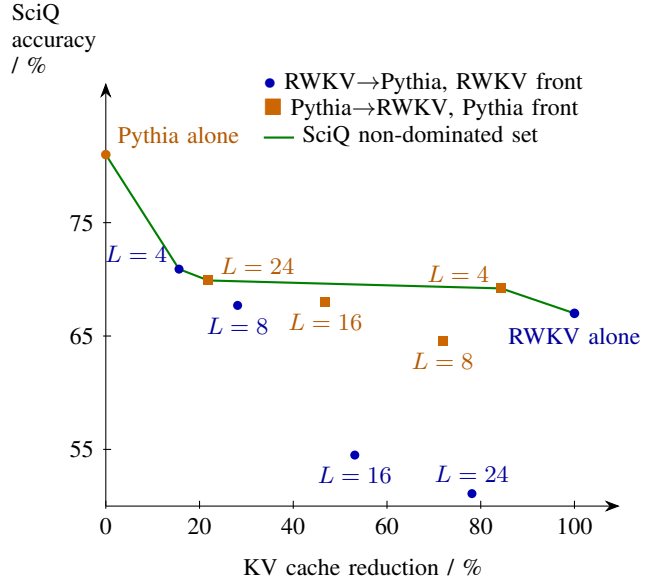
\begin{figure}[!htb]\centering
\begin{tikzpicture}[font=\small,>={Stealth[length=2mm]}]
  \def\xs{0.62}
  \def\ys{0.15}
  \draw[->] (0,0) -- (68mm,0);
  \draw[->] (0,0) -- (0,5.6) node[above left,align=left] {SciQ\\accuracy\\/ \%};
  \node at (34mm,-8mm) {KV cache reduction / \%};
  \foreach \x in {0,20,40,60,80,100} \draw (\x*\xs/10,-1pt) node[below]{\x} -- (\x*\xs/10,2pt);
  \foreach \y in {55,65,75} \draw (-1pt,{(\y-50)*\ys}) node[left]{\y} -- (2pt,{(\y-50)*\ys});
  \coordinate (pP) at (0,{(81.0-50)*\ys});
  \coordinate (ab4) at ({15.625*\xs/10},{(70.9-50)*\ys});
  \coordinate (ab8) at ({28.125*\xs/10},{(67.7-50)*\ys});
  \coordinate (ab16) at ({53.125*\xs/10},{(54.5-50)*\ys});
  \coordinate (ab24) at ({78.125*\xs/10},{(51.1-50)*\ys});
  \coordinate (ba4) at ({84.375*\xs/10},{(69.2-50)*\ys});
  \coordinate (ba8) at ({71.875*\xs/10},{(64.6-50)*\ys});
  \coordinate (ba16) at ({46.875*\xs/10},{(68.0-50)*\ys});
  \coordinate (ba24) at ({21.875*\xs/10},{(69.9-50)*\ys});
  \coordinate (rw) at ({100*\xs/10},{(67.0-50)*\ys});
  \draw[green!50!black,thick] (pP)--(ab4)--(ba24)--(ba4)--(rw);
  \fill[orange!80!black] (pP) circle (1.8pt);
  \foreach \p in {ab4,ab8,ab16,ab24} \fill[blue!70!black] (\p) circle (1.7pt);
  \foreach \p in {ba4,ba8,ba16,ba24}
    \node[fill=orange!80!black,rectangle,minimum size=3.4pt,inner sep=0pt] at (\p) {};
  \fill[blue!70!black] (rw) circle (1.8pt);
  \node[above right=-1pt and 1pt of pP,text=orange!70!black] {Pythia alone};
  \node[above left=-1pt and 1pt of ab4,text=blue!60!black] {$L=4$};
  \node[below=1pt of ab8,text=blue!60!black] {$L=8$};
  \node[below=1pt of ab16,text=blue!60!black] {$L=16$};
  \node[above=1pt of ab24,text=blue!60!black] {$L=24$};
  \node[above left=-1pt and 1pt of ba4,text=orange!70!black] {$L=4$};
  \node[below=1pt of ba8,text=orange!70!black] {$L=8$};
  \node[below=1pt of ba16,text=orange!70!black] {$L=16$};
  \node[above right=-1pt and 1pt of ba24,text=orange!70!black] {$L=24$};
  \node[below=2pt of rw,inner xsep=0pt,text=blue!60!black] {RWKV alone};
  \node[align=left,fill=white,rounded corners=1pt,inner sep=2pt] at (42mm,5.25)
    {\textcolor{blue!70!black}{$\bullet$} \RtoP{}, RWKV front\\
     \textcolor{orange!80!black}{$\blacksquare$} \PtoR{}, Pythia front\\
     \textcolor{green!50!black}{\rule[0.3ex]{4mm}{0.8pt}} SciQ non-dominated set};
\end{tikzpicture}
\caption{SciQ accuracy and KV cache reduction of \RtoP{} and \PtoR{}. The green line connects the configurations in the non-dominated set.}
\label{fig:tradeoff}
\end{figure}

Comparing \RtoP{} and \PtoR{}, \PtoRL{4} reduces the Transformer layers from $27\to5$ relative to \RtoPL{4} while keeping the two-task
mean accuracy at an equivalent level, $65.25\%\to65.70\%$. Because the ranking depends on the task, \figref{fig:tradeoff} shows the
trade-off between KV cache reduction and accuracy on SciQ. We call the non-dominated set the set of configurations for which no other configuration is at least
as good in both measures and better in at least one of them.

Adding the configurations with Pythia as the front model yields \PtoRL{4}, which has a high reduction rate and cannot be obtained from
the configurations with RWKV as the front model alone.
RWKV alone, on the other hand, needs no KV cache, and its accuracy
does not differ significantly from that of \PtoRL{4} (\S\ref{sec:goal}). The advantage of the method is therefore that it creates new trade-off points between
frozen parent models, not that new configurations uniformly exceed the parent models.

If the only goal is to reduce the KV cache, the same reduction is also obtained by running only the first blocks of Pythia and
connecting their output to the final layer normalization and the LM head. We call this method early exit. We therefore compared the accuracy of each chimera model with that of an early exit
that keeps the same number of Transformer blocks (\tabref{tab:earlyexit}). In the $5$ configurations that reduce the Transformer
layers to $17$ or fewer, the chimera model significantly exceeds the early exit on both tasks. \PtoRL{4} scores $62.2\%$ against
$26.8\%$ on ARC and $69.2\%$ against $30.5\%$ on SciQ, compared with the early exit that stops after the first $5$ blocks of Pythia.
In the $3$ configurations that keep $23$ or more layers, by contrast, the differences are small; only ARC for \RtoPL{8}, which keeps
$23$ layers, is significant, and \RtoPL{4}, which keeps $27$ layers, does not differ from the early exit.
However, because the early exit runs fewer blocks and keeps fewer resident weights at the same KV cache size, this comparison
does not equalize computation.

\begin{table}[!t]\centering\small
\caption{Differences in length-normalized accuracy (points, chimera model minus early exit) against a Pythia early exit that keeps the same number of Transformer layers, with McNemar exact-test $p$-values. Transformer: number of Transformer layers kept. R$\to$P: RWKV$\to$Pythia; P$\to$R: Pythia$\to$RWKV.}\label{tab:earlyexit}
\setlength{\tabcolsep}{3pt}
\begin{tabular}{lrrrrrr}\toprule
& & & \multicolumn{2}{c}{ARC} & \multicolumn{2}{c}{SciQ}\\
\cmidrule(lr){4-5}\cmidrule(lr){6-7}
Config. & $L$ & \shortstack{Trans-\\former} & Diff. & $p$ & Diff. & $p$\\\midrule
P$\to$R & 4  & 5  & $+35.4$ & $<10^{-4}$ & $+38.7$ & $<10^{-4}$\\
R$\to$P & 24 & 7  & $+18.5$ & $<10^{-4}$ & $+18.7$ & $<10^{-4}$\\
P$\to$R & 8  & 9  & $+27.6$ & $<10^{-4}$ & $+31.6$ & $<10^{-4}$\\
R$\to$P & 16 & 15 & $+11.2$ & $<10^{-4}$ & $+12.8$ & $<10^{-4}$\\
P$\to$R & 16 & 17 & $+16.9$ & $<10^{-4}$ & $+17.9$ & $<10^{-4}$\\
R$\to$P & 8  & 23 & $+4.9$ & $<10^{-4}$ & $+1.6$ & $0.36$\\
P$\to$R & 24 & 25 & $+2.1$ & $0.069$ & $+0.5$ & $0.80$\\
R$\to$P & 4  & 27 & $-0.2$ & $0.87$ & $-0.9$ & $0.59$\\\bottomrule
\end{tabular}
\end{table}

As for inference time, with the implementation in the Hugging Face Transformers library used in this study and the two parent models in FP16
placed on $2$ NVIDIA V100 GPUs, we compared the time to process the whole input sequence at once, taking the median of $3$ measurements. At context length $512$, RWKV alone took $6{,}538$ ms and Pythia alone
$99$ ms, a factor of about $66$. This gap arises because the implementation used runs the RWKV recurrence
sequentially, token by token. At context length $2048$ the times were $25{,}933$ ms and $385$ ms, a factor of about $67$.
For chimera models with the unused layers removed, configurations of the same direction take longer the more RWKV blocks they run;
at context length $512$, \RtoPL{4},
which runs $5$ RWKV blocks, took $1{,}128$ ms, and \PtoRL{4}, which runs $27$ RWKV blocks, took $5{,}652$ ms.
These values depend on the implementation used for the measurement and do not compare the computational cost of the architectures.

\subsection{Effect of Domain Shift}\label{sec:domain}
\begin{figure}[!htb]\centering
\begin{tikzpicture}[font=\small,xscale=0.9]
  \def\Y#1{ {10*ln(#1)/ln(10)/10} }
  \draw[->] (0,0) -- (0,3.1) node[above right] {perplexity / log scale};
  \foreach \v in {3,10,30,100,300} \draw[gray!30] (0,\Y{\v}) node[left,black]{\v} -- (84mm,\Y{\v});
  \draw[thick] (0,0) -- (84mm,0);   
  \def\grp#1#2#3#4{  
    \fill[teal!60] (#1mm-4.2mm,0) rectangle (#1mm-0.4mm,\Y{#3});
    \fill[red!40]  (#1mm+0.4mm,0) rectangle (#1mm+4.2mm,\Y{#4});
    \node[below=1pt] at (#1mm,0) {#2};
  }
  \grp{11}{RWKV}{3.42}{11.33}
  \grp{33}{Pythia}{5.19}{16.22}
  \grp{55}{\shortstack{RWKV$\to$\\Pythia\\$L=4$}}{6.40}{93.87}
  \grp{77}{\shortstack{Pythia$\to$\\RWKV\\$L=4$}}{3.85}{45.00}
  \fill[teal!60] (3mm,2.75) rectangle (7mm,2.95);
  \node[anchor=west] at (7mm,2.85) {in-domain Alpaca};
  \fill[red!40]  (36mm,2.75) rectangle (40mm,2.95);
  \node[anchor=west] at (40mm,2.85) {out-of-domain WikiText};
\end{tikzpicture}
\caption{Perplexity on Alpaca and WikiText at context length $512$. The vertical axis is logarithmic.}
\label{fig:domain}
\end{figure}
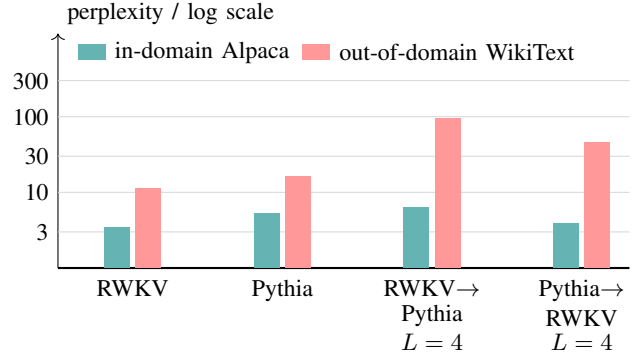

This subsection examines how performance changes when the evaluated text comes from a domain different from the training domain, a
mismatch called domain shift below.
The accuracies above were measured on short questions close to instruction format. Because a smaller KV cache matters for long
contexts, we vary the context length and measure language-modeling performance on Alpaca, inside the training domain, and on
WikiText, outside it.
As the measure we use perplexity, the exponential of the next-token cross-entropy, which is better when smaller.
\figref{fig:domain} shows the results at context length $512$. The perplexities of the parent models are $3.42$ on Alpaca and $11.33$
on WikiText for RWKV, and $5.19$ and $16.22$ for Pythia. Alpaca, however, overlaps the instruction data of both parents, so
the parents' Alpaca perplexities may partly reflect exposure during instruction tuning.
\RtoPL{4} gives $6.40$ on Alpaca and $93.87$ on WikiText, and \PtoRL{4} gives $3.85$ and $45.00$; \PtoRL{4} is lower than \RtoPL{4} in
both domains, but on WikiText it is well above those of the parent models.
Extending the context length to $2048$ gives $3.50$ on Alpaca and $34.78$ on WikiText for \PtoRL{4}, so performance does not degrade
and even improves.
The main cause of the degradation is therefore domain shift, not context length. \PtoR{} is less affected by domain shift than \RtoP{},
but the WikiText perplexity of $34.78$ at context length $2048$ is $4.2$ times the $8.27$ of RWKV, the parent model with the lower perplexity at the same
context length, and remains high.

We also compared the adapter with a control that places at the switch layer an affine map fitted by ridge regression with an
intercept, as in (iii) of \S\ref{sec:linear}. This control is refitted for each switch layer and each evaluated domain on its first
$39{,}936$ tokens, which are not used for evaluation; it is thus fitted even to WikiText, on which the adapter was not trained, so this comparison
is unfavorable to the adapter. For \PtoR{}, the WikiText perplexity of the affine control is lower than that of the adapter at every switch
layer and context length; for \PtoRL{4} it is $21.73$ against $45.00$ at context length $512$ and $15.74$ against $34.78$ at context
length $2048$. On Alpaca the two are nearly equal, $3.86$ against $3.85$ for \PtoRL{4} at context length $512$. For \RtoP{}, in
contrast, the WikiText perplexity of the affine control is at least that of the adapter at every switch layer and context length.
With this caveat, the result that the affine control is better for \PtoR{} suggests that part of the degradation outside the training domain may arise from the conversion by the adapter,
which was trained only on Alpaca.

\subsection{Generation Examples of the Chimera Models}\label{sec:chimera}
The adapter of the checkpoint at iteration $415{,}000$ was inserted at $L{=}4$ in both \RtoP{} and \PtoR{}, and responses to the same four English prompts
were generated by argmax decoding under the conditions of \S\ref{sec:common-mode}. Two of the prompts use instruction format, which
separates the instruction and the response with the headings ``\#\#\# Instruction:'' and ``\#\#\# Response:'', and two use completion
format, which asks the model to continue a sentence. $L{=}4$ is the switch layer at which the two-task mean accuracy is highest among
the four switch layers in each direction, and it is the common switch layer at which both directions belong to the non-dominated set in
the plane of KV cache reduction and accuracy on SciQ (\figref{fig:tradeoff}).
This choice, however, was made after seeing the results of the evaluation tasks.

\begin{table*}[!t]\centering\small
\setlength{\tabcolsep}{3pt}
\caption{Generation examples of the two parent models and of \RtoP{} and \PtoR{} at $L{=}4$.}\label{tab:generation}
\begin{tabular}{|P{0.175\linewidth}|P{0.185\linewidth}|P{0.2\linewidth}|P{0.175\linewidth}|P{0.195\linewidth}|}\hline
Prompt (verbatim) & Pythia alone &
\RtoPL{4} &
RWKV alone &
\PtoRL{4}\\\hline
{\ttfamily \#\#\# Instruction:\par
What is the capital of France?\par\smallskip
\#\#\# Response:} &
``The capital of France is Paris.'' & ``The capital of France is Paris.'' \checkmark &
``Paris.'' & ``Paris.'' \checkmark\\\hline
{\ttfamily The three primary colors are} &
``red, yellow, and blue. These colors are often used in combination to create a wide variety of colors.'' & ``red, yellow, and blue.'' \checkmark &
``red, blue, and yellow.'' then drifts into a Q\&A asking for green, the opposite color of red &
``red, blue, and yellow.'' then drifts into a Bob/Alice dialogue about green\\\hline
{\ttfamily Question: Water is made of hydrogen and\par Answer:} &
``Hydrogen is made of hydrogen and oxygen.'' &
``Hydrogen.'' wrong &
``hydrogen.'' wrong & ``hydrogen.'' wrong\\\hline
{\ttfamily \#\#\# Instruction:\par
Explain what a black hole is in one sentence.\par\smallskip
\#\#\# Response:} &
``A black hole is a region of spacetime where gravity is so strong that nothing, not even light, can escape its pull.'' &
``A black hole is a region in space where gravity is so strong that nothing, not even light, can escape its pull.'' \checkmark &
``A black hole is a region of space where the gravitational pull is so strong that nothing, not even light, can escape.'' &
``A black hole is a type of celestial object that is formed when a star collapses and collapses into a black hole.''
circular\\\hline
\end{tabular}
\end{table*}

These examples show that both \RtoP{} and \PtoR{} can generate syntactically well-formed output. Both directions answer the capital of
France correctly and begin listing the primary colors correctly. \RtoP{} also answers the black-hole instruction correctly in one
sentence, so generation that follows the instruction is achieved. On the other hand, content or instruction format can be lost even when
fluency is kept. Both directions answer ``hydrogen'' wrongly in the completion about the composition of water, and \PtoR{} drifts into a
dialogue after answering the primary colors correctly and explains the black hole circularly.
Current chimera models can therefore lack factual accuracy. RWKV alone, however, also gives the wrong answer about water and drifts
after the primary colors, and a causal relation between the conversion error at $L{=}4$ and individual wrong answers has not been verified.

\section{Discussion}\label{sec:discussion}
In this experiment, a shared representation with $\rho_{\mathrm{ctr}}\!=\!0.901$ could be learned between corresponding layers of frozen
models from different families. The range in which this conclusion applies, however, is narrow. The correspondence was measured under
the favorable conditions of the same tokenizer, hidden width, depth, and input, in a domain close to Alpaca. The shuffle control rules out a
token-independent match of distributions, but we have not separated whether the observed correspondence arises from meaning, vocabulary,
position, or syntax. Performance also decreases sharply on WikiText, outside the training domain (\S\ref{sec:domain}).
This study therefore claims only the existence of a shared representation that corresponds token by token under favorable conditions,
not a general common semantic space.

A successful functional stitch does not mean that the two models represent the same information. Smith et al.~\cite{pmlr-v267-smith25a}
showed that stitching can succeed between different tasks, between images and bird-song audio, and
even from randomly generated, clustered pseudo-representations. The stitching layers of their discriminative models, however, were
trained with the task loss, the scheme that Balogh and Jelasity~\cite{balogh2025not} point out rates distant layers as similar by
producing out-of-distribution representations. Our adapter, which uses no task loss and is trained with self-reconstruction and latent
alignment, is closer to the direct matching of representations that the same paper finds to balance structural and functional
requirements. Smith et al. nonetheless obtained apparent alignment even with low-capacity stitching layers and with autoencoder stitches
trained by matching latent codes, and our adapter has higher capacity. Our main reason for judging that the observed
alignment cannot be explained by the capacity of the adapter alone is that a low-capacity affine map already achieves substantial
cross-family prediction (\S\ref{sec:linear}).

A high latent correlation also does not guarantee high read-out accuracy, so alignment and decodability must be evaluated separately,
and other work further distinguishes decodability from use by the receiving model. The preprint of Zhang and Xin~\cite{zhang2026negativeresultcrossmodelactivation}
showed that a linear map that transfers activations between small models of the Pythia series achieves $R^2=0.883$ in a normalized space, yet replacing the
hidden states of the receiving model with the converted activations lowers performance on tasks that require several reasoning steps,
even after rescaling to the receiver's norm, and weak additive injection does not improve it. They concluded that the training objective should optimize how the receiving
model uses the activations; this point also applies to our adapter, whose objective does not include the receiving model's output.
The unstandardization in Equation~\eqref{eq:boundary} differs from their norm rescaling in restoring the per-dimension mean and scale,
but we have not isolated the contribution of this difference.

\section{Limitations of This Study}\label{sec:limitations}
Some of the evaluation questions may be contained in the training data of a parent model. Pythia was instruction-tuned on a subset of
FLAN V2, and the tasks that make up FLAN V2 include ARC and SciQ. Matching the corresponding part of the released mixture verbatim finds
$19$ ARC questions and $4$ SciQ questions among the evaluation questions. The matching covers only verbatim matches and cannot detect
paraphrases or formatting differences. Questions shorter than $6$ words, which we excluded to avoid chance matches, were not searched
either, $40$ for ARC and $11$ for SciQ. The detected counts are therefore lower bounds. The matching also covers only the FLAN V2 and Chain-of-Thought parts of Pythia's mixture,
not RWKV's instruction data or The Pile, on which both models were pretrained. Recomputing the accuracy of every configuration
from the per-question correctness records without these questions changes it by at most $0.22$ points and does not change the
conclusion of any comparison discussed in the text. On these questions, moreover, the accuracy of RWKV, whose listed instruction data do not include FLAN V2,
changes relative to the remaining questions in the same direction as that of Pythia, by $-9.2$ points on ARC and $+8.0$ points on SciQ, so this pattern is also consistent with an explanation based on the difficulty of the
questions themselves. ARC and SciQ nonetheless remain kinds of tasks that Pythia has
already learned, so the values in this paper should be read as measuring the retention of ability after composition rather than the
acquisition of new ability.

The adapter was trained with a single random seed, and although $2$ tasks$\times2$ directions$\times4$ switch layers and several
controls were evaluated, no correction for multiple comparisons was applied. The main read-out accuracies were also computed on
evaluation data that were used to adjust the learning rate and to select the checkpoint, and may be somewhat optimistic. The same
caution applies to the values in \S\ref{sec:linear}, which use sequences overlapping these evaluation data.
The intervention on the nonlinear part was performed on a single checkpoint, and whether it generalizes to adapters trained under
different conditions is unknown.
Each composition also has a single switch layer, and how conversion error accumulates over several switch layers has not been
measured. Furthermore, the multiple-choice evaluation is based on likelihood and does not directly measure the ability to generate the correct
answer, and the four generation examples in each direction are qualitative supplements, not a comprehensive evaluation of generation
ability.

The long-context cross-entropy was measured up to context length $2048$ to match the maximum number of positions in Pythia's configuration.
Measurement points beyond the $1024$ specified by the converted RWKV configuration lie outside the range the configuration assumes, and,
as noted in \S\ref{sec:method}, the context length used to train the original checkpoint cannot be confirmed.
This evaluation also measures language-modeling quality and does not directly measure reading comprehension that requires long-range
dependencies. Ablations over $\sigma$, $\lambda$, the latent width, the number of residual blocks, and the amount of training data, as
well as an alignment control with two models of the same family, were not carried out in this study and are left
for future work.

\section{Conclusion}\label{sec:conclusion}
We built NinaXander, a series of chimera models that connect frozen blocks from different architecture families at a single point
through one shared-latent adapter, and tested with RWKV and Pythia whether frozen models from different families
can be recombined post hoc. Under favorable conditions the alignment between corresponding layers is learnable, and the $8$
chimera models over $2$ directions $\times$ $4$ switch layers, obtained from one adapter without retraining, answered multiple-choice
questions, and the $2$ configurations at $L{=}4$ produced syntactically well-formed text. \PtoRL{4} reduced the Transformer KV cache by $84.4\%$ with accuracy not significantly different from that of RWKV alone,
but it did not exceed Pythia. The configurations that exceed a Pythia early exit with the same KV cache size by a wide margin on both
tasks, however, are those that reduce the Transformer layers to $17$ or fewer. In all layers, each cross-family read-out was also lower than the
self-reconstruction that uses the same decoder, and performance decreased sharply on WikiText, outside the training domain. The significance of NinaXander therefore lies not in a
higher-performing new model but in showing a way to recombine frozen models post hoc. The contribution of this study is to show that this
recombination is partly possible
and can create new trade-off points between frozen parent models, and to describe, on the basis of measurements, the possibilities and
limits of composition in one favorable case. Future work includes generalization through training across several
domains, identifying what limits the self-reconstruction accuracy, selecting complementary parent models, measuring how conversion
error accumulates over several switch layers, and reproduction with multiple
random seeds.

\section*{Acknowledgment}
This work was supported by the JST Next-Generation Edge AI Semiconductor Research and Development Program, Grant Number JPMJES2511.
This work was also supported by the Joint Usage/Research Center for Interdisciplinary Large-scale Information Infrastructures (JHPCN) in Japan (Project ID: jh260017).

\bibliographystyle{IEEEtran}
\bibliography{reference}

\appendix
\section{Reproduction Information}\label{sec:repro}
\begin{itemize}\setlength{\itemsep}{1pt}\raggedright
\item Models: the two parent models of \S\ref{sec:method} are used in FP16. The released artifact
\texttt{parent\_fingerprints.json} records the SHA-256 (Secure Hash Algorithm 256-bit) hashes of the local parent models used for training.
\item Adapter training: the adapter was trained on $4$ NVIDIA V100 GPUs.
\item Data: the evaluation sequences were taken from Alpaca from character position $2{,}000{,}000$ onward, which does not depend on
the number of training tokens. The standardization statistics are computed from the training residuals of each layer
and are also used for unstandardization.
\item Evaluation: for the RWKV back model of \PtoR{}, options are always evaluated with a batch size of $1$.
\item Code and structured data: the repository given in \S\ref{sec:method} contains the experiment code, the raw data of each
evaluation with aggregate tables of the reported values, the reproduction procedures, and a check script that automatically matches the numbers in the text against the aggregate tables.
\item Hugging Face releases: the inference artifacts as of July 30, 2026 are stored in the following three model repositories.
\begin{itemize}\setlength{\itemsep}{0pt}\raggedright
\item Shared adapter: \hfrepo
  {https://huggingface.co/Katagiri-Hoshino-Lab/ninaxander-raven7b-tulu69-adapter}
  {ninaxander-raven7b-tulu69-adapter}
\item \PtoR{} with the default $L{=}4$: \hfrepo
  {https://huggingface.co/Katagiri-Hoshino-Lab/ninaxander-tulu69-to-raven7b-best}
  {ninaxander-tulu69-to-raven7b-best}
\item \RtoP{} with the default $L{=}4$: \hfrepo
  {https://huggingface.co/Katagiri-Hoshino-Lab/ninaxander-raven7b-to-tulu69-best}
  {ninaxander-raven7b-to-tulu69-best}
\end{itemize}
The latter two bundle the two frozen parent models in FP16, the shared tokenizer, the adapter, and custom inference code that selects the switch
layer from $L{=}4,8,16,24$, so each runs as a standalone chimera model. Each is fixed to one connection direction, and its default switch layer
$L{=}4$ was chosen after seeing the evaluation results, as in \S\ref{sec:chimera}. No independent data split was used for this selection.
All three repositories contain model cards in English, Japanese, and Simplified Chinese, a NOTICE file that records the attribution
of the parent models, weights in the Safetensors format, and a file list with SHA-256 hashes. The license is the AI2 AI Model License
for non-commercial research, matching Pythia, the more restrictive parent model.
\end{itemize}

\end{document}